\documentclass[letterpaper, 10 pt, conference]{ieeeconf}
\IEEEoverridecommandlockouts
\usepackage{graphicx} 
\usepackage{subcaption} 
\usepackage{xcolor} 
\usepackage{amsmath} 

\usepackage{colortbl}
\usepackage{booktabs}
\usepackage{tabularx}

\makeatletter
\let\NAT@parse\undefined
\makeatother
\usepackage[hidelinks]{hyperref} 
\usepackage{url}

\begin{document}
\title{\LARGE \bf A Dynamic Points Removal Benchmark in Point Cloud Maps}
\author{Qingwen~Zhang$^{1}$, Daniel~Duberg$^{1}$, Ruoyu~Geng$^{2}$, Mingkai~Jia$^{2}$, Lujia~Wang$^{2}$, Patric~Jensfelt$^{1}$
\thanks{$^{1}$Authors are with the Division of Robotics, Perception, and Learning (RPL), KTH Royal Institute of Technology, Stockholm 114 28, Sweden. (email: qingwen@kth.se)}
\thanks{$^{2}$Authors are with Robotics Institute, The Hong Kong University of Science and Technology, Hong Kong SAR, China.}
}
\maketitle

\begin{abstract}
In the field of robotics, the point cloud has become an essential map representation.
From the perspective of downstream tasks like localization and global path planning, points corresponding to dynamic objects will adversely affect their performance.
Existing methods for removing dynamic points in point clouds often lack clarity in comparative evaluations and comprehensive analysis. 
Therefore, we propose an easy-to-extend unified benchmarking framework for evaluating  techniques for removing dynamic points in maps.
It includes refactored state-of-art methods and novel metrics to analyze the limitations of these approaches. 
This enables researchers to dive deep into the underlying reasons behind these limitations. 
The benchmark makes use of several datasets with different sensor types.
All the code and datasets related to our study are publicly available for further development and utilization.
\end{abstract}

\section{Introduction}
\begin{figure}[t]
\centering
\includegraphics[width=3.3in]{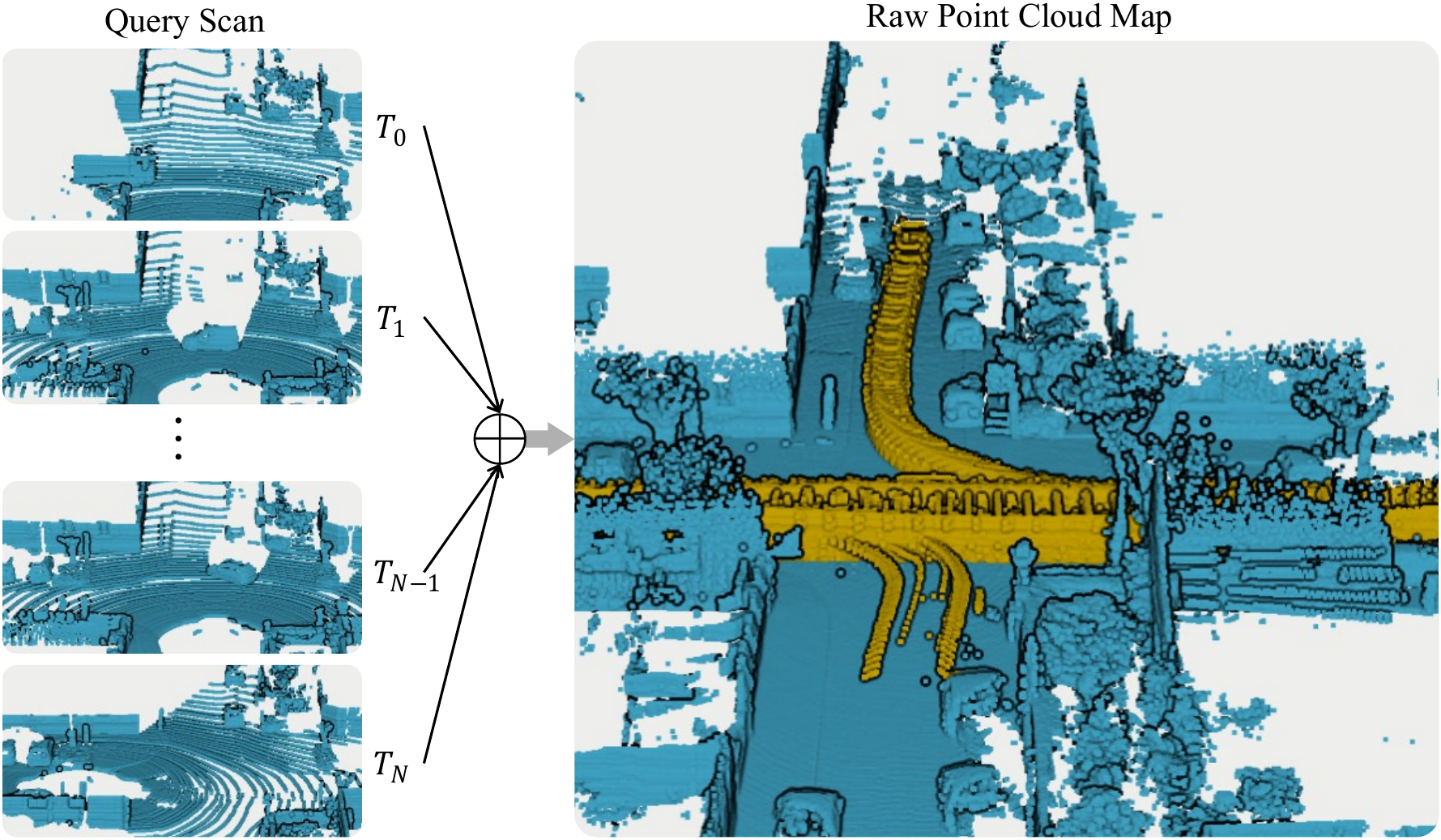}
\caption{Illustration of ghost points resulting from dynamic objects in KITTI sequence 7. The yellow points to the right represent points labeled as belonging to dynamic objects in the dataset. These ghost points negatively affect downstream tasks and overall point cloud quality.}
\label{fig:backgroud}
\vspace{-1.0em} 
\end{figure}

Point clouds are widely used in the domains of robotics, given their effectiveness in facilitating key components such as localization and path planning. Current SLAM (Simultaneous Localization and Mapping) packages \cite{ TMSLICT, jjhaoMLOAM, xu2022fast} fuse data from multiple sensors to obtain corresponding poses. These poses can be used to integrate point cloud frames into a global map shown in Fig.~\ref{fig:backgroud}.

Removing dynamic points from maps is crucial for accurate representations of the environment. Failing to detect dynamic points while integrating point cloud data can result in the inclusion of ghost points, as illustrated in Fig.~\ref{fig:backgroud} yellow part. 
In the localization task, ghost points may reduce robustness as they introduce ambiguous features or mislead the matching process between the current observation and the global map. For global path planning, the presence of ghost points can lead to suboptimal path selection. 
If the planning algorithm interprets points corresponding to dynamics as part of the static environment's structure, it will mistake these points as obstacles and classify the region as untraversable, resulting in unnecessarily long path allocation or even failure in path planning.

Various methods are proposed to tackle the issue of removing dynamic points, where different metrics are tailored to showcase the benefits of their own approaches. For example, Lim \textit{et al.}~\cite{lim2021erasor} utilize the voxel-wise preservation rate to evaluate their results. However, the existing evaluation metrics neglect the classification accuracy in the sub-voxel scale.
Our benchmark adopts a set of new metrics for point-wise evaluation with a unified standard.

Existing methods including~\cite{lim2021erasor, gskim-2020-iros} are mainly evaluated on SemanticKITTI~\cite{behley2019iccv}, which solely includes small town scenarios by a single type of LiDAR. 
Our benchmark performs evaluation on various datasets to analyze robustness towards different scenarios and sensor setups.
We also prepared a dataset in a semi-indoor scenario where dynamic objects are moving close to the static structure, and the ego agent is equipped with a sparse LiDAR.
For qualitative results, we additionally choose the latest Argoverse 2.0 dataset~\cite{Argoverse2} that contains various streetscapes in big cities and has more dynamic objects compared with SemanticKITTI.
These diverse datasets enable a comprehensive  assessment of the existing techniques to compare their adaptability to a range of scenarios and sensor configurations.

Based on our benchmarking result, we summarise the strengths and weaknesses of each technique revealed from our proposed metrics, facilitating further development and innovation in the field.
For instance, the occupancy mapping approach, Octomap~\cite{hornung13auro}, is also frequently used as a dynamic point removal baseline. 
Guided by the benchmarking result, we demonstrate how we improve its dynamic point removal performance by incorporating ground fitting into the pipeline.

We contribute the benchmark implementation and extended datasets to the research community at \color{blue}~\url{https://github.com/KTH-RPL/DynamicMap_Benchmark}.\color{black}

The main additional contributions include the following:
\begin{itemize}
\item Refactoring existing methods to establish a unified benchmark for removing dynamic points in the map. 
\item Introducing new metrics and evaluating the performance of all methods, detailing the challenges associated with this task.
\item Introducing an extension of Octomap better adapted to the map clean task.
\end{itemize}

\section{Related work}
\label{sec:related_work}
In the field of point cloud processing on dynamic points removal, methods can be broadly categorized into two main approaches: learning-based and traditional algorithms. 
Learning-based methods have been increasingly popular in detecting dynamic points or objects. However, they require training data and a network to learn latent space representations, often lacking explainability. Therefore, this paper focuses on traditional approaches to removing dynamic points. In the below sections, we will review both learning-based and traditional methods in detail.

\subsection{Learning-based}
Learning-based methods typically involve deep neural networks and supervised training with labeled datasets. 
Mersch \textit{et al.}~\cite{9796597} employ sparse 4D convolutions to segment receding moving objects in 3D LiDAR data, efficiently processing spatiotemporal information using sparse convolutions. 
Sun \textit{et al.}~\cite{9981210} develop a novel framework for fusing spatial and temporal data from LiDAR sensors, leveraging range and residual images as input to the network. Toyungyernsub \textit{et al.}~\cite{9981323} predict urban environment occupancy by considering both spatial and temporal information, incorporating environmental dynamics to improve moving object segmentation performance. 
Huang \textit{et al.}~\cite{huang2022dynamic} propose a novel method for unsupervised point cloud segmentation by jointly learning the latent space representation and a clustering algorithm using a variational autoencoder. 
Lastly, Khurana \textit{et al.}~\cite{khurana2022differentiable} use differentiable raycasting to render future occupancy predictions into future LiDAR sweep predictions for learning, allowing geometric occupancy maps to clear the motion of the environment from the motion of the ego-vehicle.

However, they share common drawbacks, such as the need for extensive labeled datasets, unbalanced data during training \cite{zhang2022not}, and potential limitations when applied to different sensor types they were not trained on.

\subsection{Traditional Algorithm}
\label{sec:related_work_traditional}
In light of these challenges, our focus shifts towards traditional methods, which typically exhibit greater robustness and flexibility in handling diverse sensor types and data distributions.
Various approaches have been proposed, often categorized into ray-casting, visibility-based, and visibility-free.

Occupancy grids, often in the form of Octomap~\cite{hornung13auro}, are popular techniques that employ ray casting to update the occupancy value of the grid map space by counting the hits and misses of scans. Additionally, other data structures have been proposed, e.g., by representing the truncated signed distance field (TSDF)~\cite{oleynikova2017voxblox} instead of occupancy. They rely on the concept of occupancy values or truncated signed distances to detect dynamic points in point clouds. These methods update the values for each voxel, frame by frame, based on the measurements obtained from the sensor. 
If the values within a voxel deviate significantly from a specified threshold, the points inside that voxel are considered dynamic.

Despite their effectiveness, these methods can be computationally expensive when performing ray-casting steps, leading to the development of visibility-based methods to reduce computational costs.
Visibility-based methods assume that if a query point is observed behind a previously acquired point in the map, then the previously acquired point is dynamic. Kim \textit{et al.} \cite{gskim-2020-iros} constructs a static point cloud map using multi-resolution range images based on visibility.

\begin{figure}[t]
    \centering
    \begin{subfigure}[b]{0.45\textwidth}
        \centering
        \includegraphics[width=\textwidth]{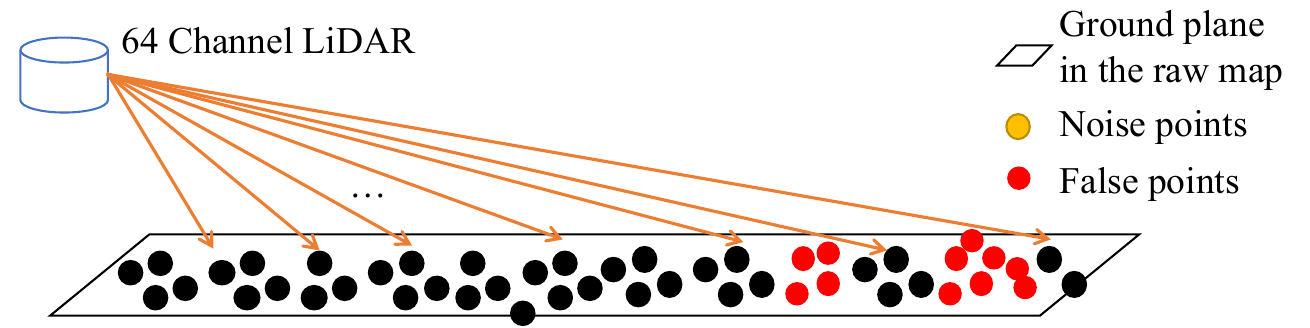}
        \vspace{-10pt}
        \caption{Ray angle challenges}
        \label{fig:issue_a}
    \end{subfigure}
    \begin{subfigure}[b]{0.45\textwidth}
        \vspace{0.5em}
        \centering
        \includegraphics[width=\textwidth]{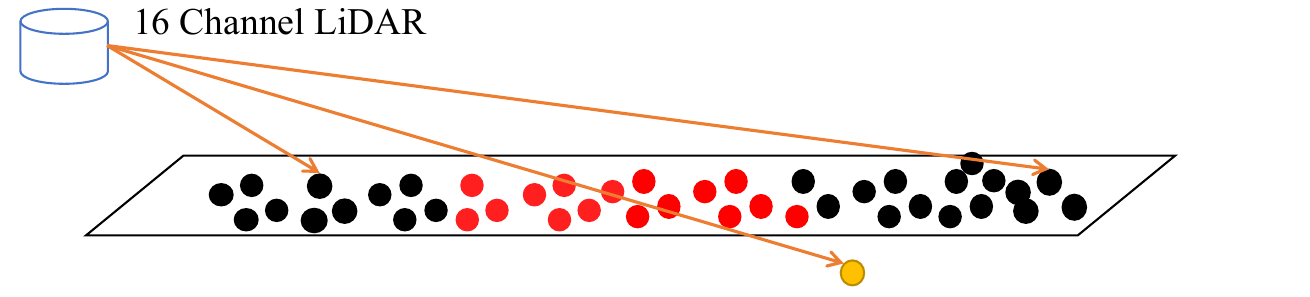}
        \vspace{-10pt}
        \caption{Sparse LiDAR queries}
        \label{fig:issue_b}
    \end{subfigure}
    \begin{subfigure}[b]{0.45\textwidth}
        \centering
        \includegraphics[width=\textwidth]{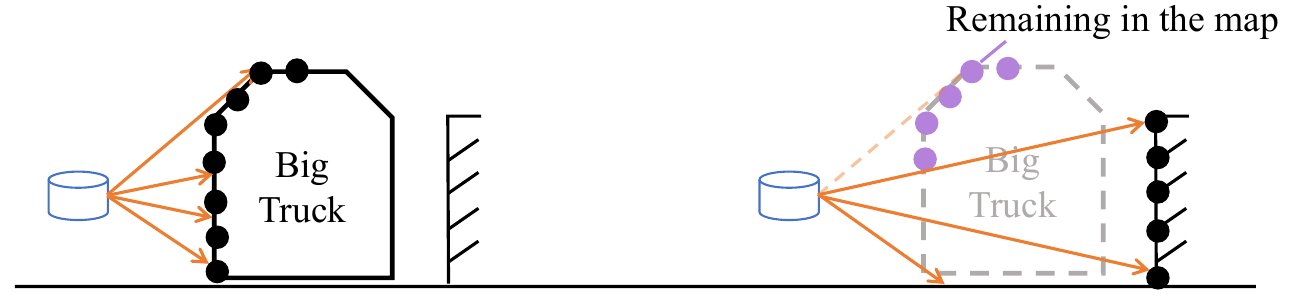}
        \vspace{-10pt}
        \caption{Non-removable dynamic points}
        \label{fig:issue_c}
    \end{subfigure}
    \caption{Limitation on ray casting-based and visibility-based methods.}
    \label{fig:issue_observe}
    \vspace{-1.0em} 
\end{figure}

Both ray casting and visibility-based methods suffer from the problems illustrated in Fig.~\ref{fig:issue_observe}. 
(a) shows that rays are far from the ground, and the angle between the rays and the ground line becomes very small. In such scenarios, ray-based methods update the free value when the rays pass through the area, which may cause some ground points to be incorrectly seen as dynamic. 
(b) means after accumulating multiple scan frames, noise below the ground in some frames can cause previous regions to be updated as free, mislabeling ground points. 
(c) illustrates how these methods fail to remove dynamic points when no object is behind them. In this example, only some hits on the big truck will later be cleared by hits on the wall, while others will not. The purple hits will erroneously remain, as no new hits pass through them.

Lim \textit{et al.} observed these limitations in \cite{9361109} and proposed a novel approach based on the height difference between the raw map and the query. 
They compare the ratio between the difference in the minimum and maximum z-values in regions between a query scan and the map. 
If the ratio is larger than a predefined threshold, the region is considered to contain dynamic objects. This approach improved the handling of dynamic objects from unlabeled classes.

\begin{figure}[t]
    \centering
    \begin{subfigure}[t]{0.8\linewidth}
        \centering
        \includegraphics[width=\textwidth]{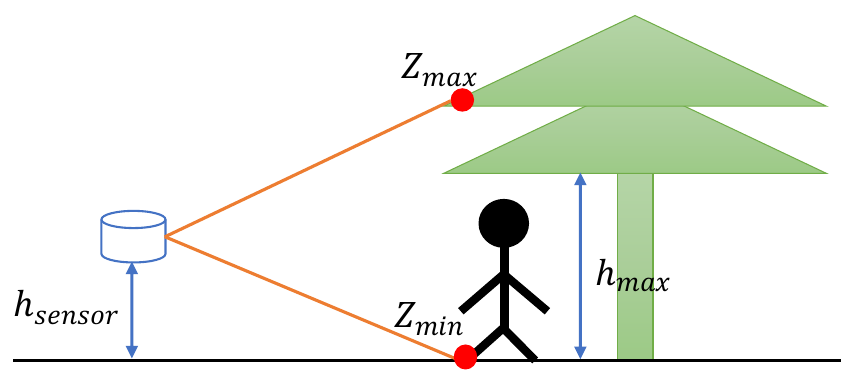}
        \caption{}
    \end{subfigure}
    \hspace{0.1\linewidth}
    \begin{subfigure}[t]{0.8\linewidth}
        \centering
        \includegraphics[width=\textwidth]{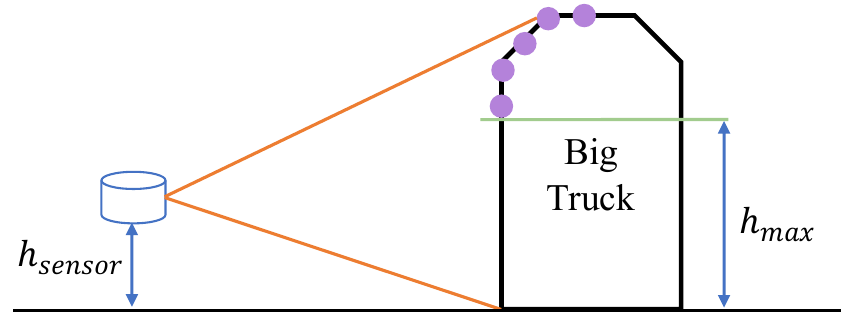}
        \caption{}
    \end{subfigure}
    \caption{Limitations of height-threshold-based methods. (a) When people stand under a tree, using a height threshold $h_{max}$ typically ignores the highest points that the sensor observed.
    (b) However, when choosing a threshold $h_{max}$, larger objects such as a truck may still have remaining points.}
    \label{fig:issue_observe_height}
    \vspace{-1.0em} 
\end{figure}

We have discussed several traditional methods for removing dynamic points from point clouds. 
They often involve numerous parameters that need to be tuned.
For instance, 
Lim's method \cite{9361109} requires knowledge of the sensor height, making it highly sensitive to height values. This approach also necessitates tuning the maximum and minimum height ranges, as it cannot handle scenarios such as pedestrians walking under trees in Fig.~\ref{fig:issue_observe_height}. 

This paper remains focused on traditional methods because they do not require the creation of a large labeled dataset or training on various datasets to ensure generalization, as compared to learning-based approaches. 
Nevertheless, it is still possible to include learning-based methods with the effort dedicated to creating various labeled datasets, training networks, and performing inference under unified setups.

\section{Methods}
\label{sec:benchmark_methods}
In this section, we provide a summary of the methods \cite{hornung13auro, gskim-2020-iros, 9361109} included in our benchmarks, discussing their algorithm design and frameworks. We prepare the processing dataset and scripts to extract data from several open-dataset and refactored methods without ROS (Robot Operating System) for easier benchmarking and faster running speeds.

Guided by our benchmark analysis and addressing the angle problem and sparse points problems in Fig.~\ref{fig:issue_observe}, we adapt Octomap~\cite{hornung13auro} to estimate the ground, followed by the same ray casting process for hit-and-miss detection in non-ground points. 

\begin{figure}[t]
    \centering
    \hspace{0.05\linewidth}
    \begin{subfigure}[t]{1.0\linewidth}
        \centering
        \includegraphics[width=\textwidth]{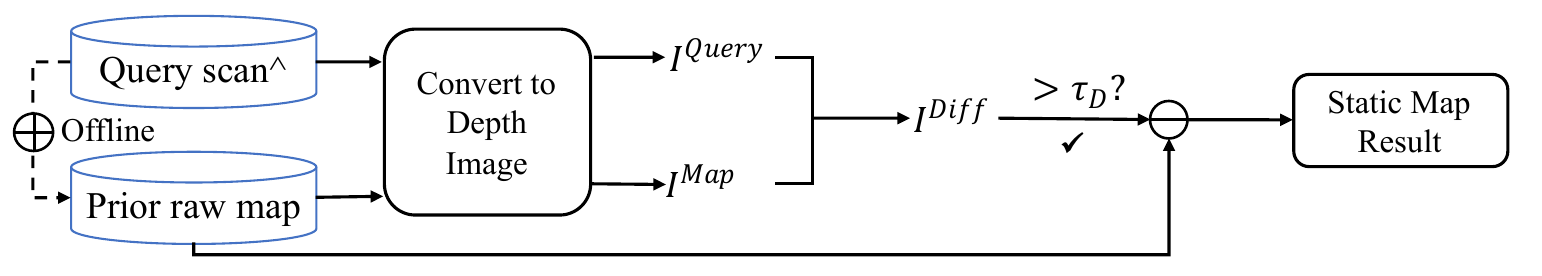}
        \caption{Removert~\cite{gskim-2020-iros}}
        \label{fig:removert}
    \end{subfigure}
    \begin{subfigure}[t]{1.0\linewidth}
        \vspace{0.5em}
        \centering
        \includegraphics[width=\textwidth]{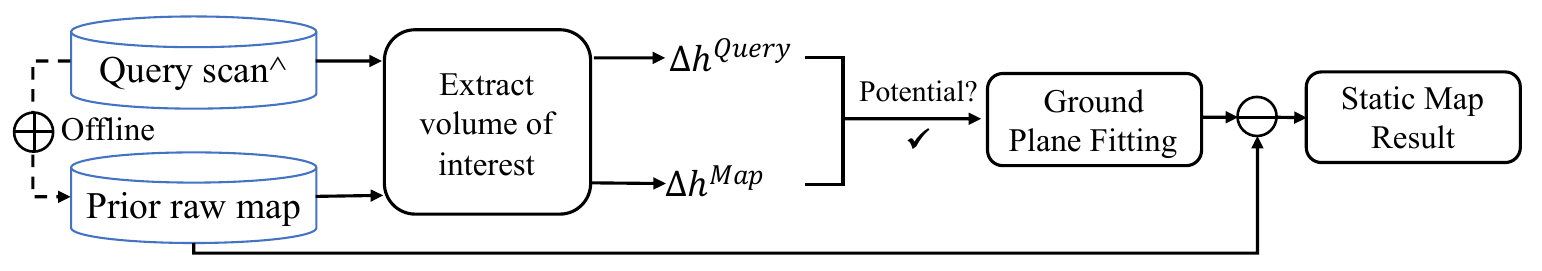}
        \caption{ERASOR~\cite{9361109}}
        \label{fig:erasor}
    \end{subfigure}
    \begin{subfigure}[t]{1.0\linewidth}
        \vspace{0.5em}
        \centering
        \includegraphics[width=\textwidth]{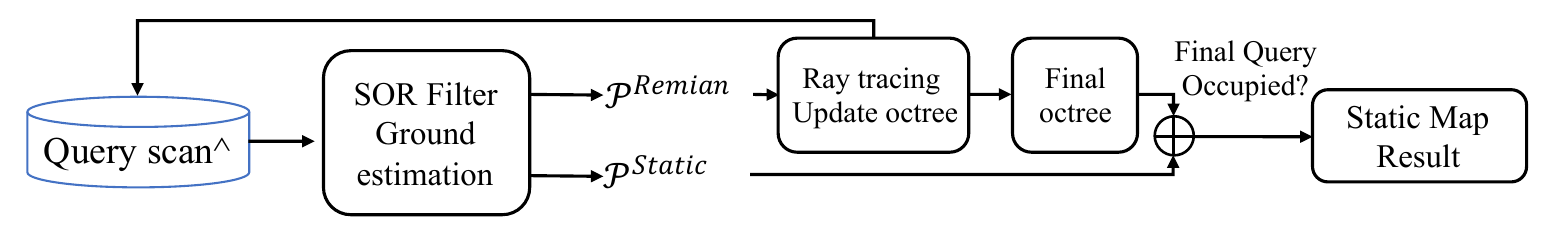}
        \caption{Improved Octomap}
        \label{fig:octomap}
    \end{subfigure}
    \caption{Framework of different methods. The term `Query Scan\^{}' indicates that the data has already been transformed to the world frame with the sensor center pose included.}
    \label{fig:framework}
    \vspace{-1.0em} 
\end{figure}
\subsection{Removert}
Kim \textit{et al.} \cite{gskim-2020-iros} proposes an offline method that requires a prior raw map to compare the difference between query and raw as shown in Fig.~\ref{fig:framework}. Firstly, they convert the query and prior raw map point cloud to depth range images using OpenCV~\cite{opencv_library}. Subsequently, they compute the difference between these two image matrices $I_k^Q, I_k^M$ as follows:
\begin{align}
I_k^{{\text{Diff}}} = I_k^Q - I_k^M.
\end{align}

Finally, the dynamic map points are defined in \cite{gskim-2020-iros}, 
\begin{align}
\mathcal{P}_k^{DM} = { {{\mathbf{p}}_{k,ij}^M \in {\mathcal{P}_k^M} | I_{k,ij}^{{\text{Diff}}} > {\tau _D}} },
\end{align}
where $\mathcal{P}_k^M$ is the raw global point cloud map, $\mathcal{P}_k^{DM}$ is the set of dynamic points, and ${\mathbf{p}}_{k,ij}^M$ is the set of points in the pixel $(i,j)$, $\tau _D$ is a threshold.
Fig.~\ref{fig:removert} illustrates their framework.

\subsection{ERASOR}
Lim \textit{et al.} \cite{9361109} propose an approach based on the observation that most dynamic objects in urban environments are in contact with the ground. They introduce the novel concept of pseudo-occupancy to represent the occupancy of unit space and discriminate spaces with varying occupancy levels. Subsequently, they determine potential dynamic candidate bins $\mathcal{P}_{Dynamic}$ based on the height difference between the raw map and query frame, as briefly described in \cite{9361109}:
\begin{align}
\Delta h_{(i,j), t} = \sup {\lbrace Z_{(i,j), t}\rbrace } - \inf {\lbrace Z_{(i,j), t}\rbrace }
\end{align}
where $Z_{(i,j), t}=\{z_k\in\mathbf{p}_{k}|\mathbf{p}_{k}\in\mathcal{S}_{(i,j), t}\}$, and $z$ represents the point's z-value concerning the sensor origin. $\sup, \inf$ separately means the highest and lowest point height value in the $Z$.

The condition for determining potential dynamic candidate bins is:
\begin{equation}
\begin{aligned}
&\text{if } \frac{\Delta h_{(i,j), t}^{Query}}{\Delta h_{(i,j), t}^{Map}} < 0.2, \\
&\text{then } \mathcal{S}{(i,j), t} \in \mathcal{P}_{Dynamic}
\end{aligned}
\end{equation}

Finally, they employ Region-wise Ground Plane Fitting (R-GPF) to distinguish static points from dynamic points within the candidate bins that potentially contain dynamic points. Fig.~\ref{fig:erasor} illustrates their framework.

\subsection{Octomap and Improvement}
\label{sec:improved_octomap}
Hornung \textit{et al.} \cite{hornung13auro} offer a popular mapping framework to generate volumetric 3D environment models in the robotics field. It is based on octrees and uses probabilistic occupancy estimation. Although it is not initially designed for dynamic point removal, it has frequently been used as a baseline.

First, Octomap rasterizes all points to 3D voxels, where each voxel is a leaf node $n$. Then, the probability of $n$ being occupied is updated given the sensor measurements $z_{1:t}$ according to:

\begin{equation}
\begin{aligned}
P&\left(n \mid z_{1: t}\right) \\
=&\left[1+\frac{1-P\left(n \mid z_t\right)}{P\left(n \mid z_t\right)} \frac{1-P\left(n \mid z_{1: t-1}\right)}{P\left(n \mid z_{1: t-1}\right)} \frac{P(n)}{1-P(n)}\right]^{-1}
\end{aligned}
\end{equation}

After updating the whole map with all scan frames, each node in the map will have a final occupancy value. If it exceeds a threshold, we consider the node as a static point. During this process, the occupancy probability of nodes containing dynamic points will decrease as rays pass through these nodes in some frames, reducing their occupancy values.

However, as mentioned earlier, it is not designed for dynamic point removal tasks, and in Section~\ref{sec:evaluation}, we can observe the challenges mentioned in Section~\ref{sec:related_work}. Guided by our benchmarking analysis, we will enhance the original Octomap by incorporating noise filtering and ground estimation techniques. The performance differences between our improved Octomap and the original version are examined through ablation studies in Section~\ref{sec:evaluation}, demonstrating the benefits of our modifications.

To minimize the impact of noise and abnormal points or reduce the computational burden of our ray casting, we employ the Statistical Outlier Removal (SOR) technique for filtering. Then we perform ground estimation using Sample Consensus (SAC) segmentation~\cite{Rusu_ICRA2011_PCL} on the output point clouds. In the last, we optimize the process by setting the grid cells occupied by the estimated ground points as free, ensuring no ray casting occurs in these regions. This approach prevents the mislabeling of ground points as dynamic and their subsequent removal from the static map, preserving the integrity of the final representation. We then only integrate and update the octree based on the non-ground points throughout all frames. 

When exporting the final map, we use a threshold to query the occupancy grid points and integrate the ground points.
This approach ensures that the occupancy values of grid cells containing dynamic points are updated when the dynamic objects move away, and rays pass through the area once more, providing an accurate and efficient representation of the environment.

\section{Benchmark Setup}
\renewcommand{\arraystretch}{1.2}
\begin{table*}[t!]
\caption{Quantitative comparison of dynamic object removal methods}
\centering
\begin{tabular}{l|ccc|ccc|ccc|ccc} 
\toprule
& \multicolumn{3}{c|}{KITTI sequence 00}
& \multicolumn{3}{c|}{KITTI sequence 05}
& \multicolumn{3}{c|}{AV2.0 big city}
& \multicolumn{3}{c}{Semi-indoor}                                                                         \\ 
\hline
Methods & \multicolumn{1}{l}{SA ↑} & \multicolumn{1}{l}{DA ↑} & \multicolumn{1}{l|}{{\cellcolor[rgb]{0.949,0.949,0.949}}AA ↑} & \multicolumn{1}{l}{SA ↑} & \multicolumn{1}{l}{DA ↑} & \multicolumn{1}{l|}{{\cellcolor[rgb]{0.949,0.949,0.949}}AA ↑} & \multicolumn{1}{l}{SA ↑} & \multicolumn{1}{l}{DA ↑} & \multicolumn{1}{l|}{{\cellcolor[rgb]{0.949,0.949,0.949}}AA ↑} & \multicolumn{1}{l}{SA↑} & \multicolumn{1}{l}{DA ↑} & \multicolumn{1}{l}{{\cellcolor[rgb]{0.949,0.949,0.949}}AA ↑}  \\ 
\hline
Removert\textsuperscript{*}~\cite{gskim-2020-iros}  & 99.44                    & 41.53                    & {\cellcolor[rgb]{0.949,0.949,0.949}}64.26                     & 99.42                    & 22.28                    & {\cellcolor[rgb]{0.949,0.949,0.949}}47.06                     & 98.97                    & 31.16                    & {\cellcolor[rgb]{0.949,0.949,0.949}}55.53                     & 99.96                    & 12.15                    & {\cellcolor[rgb]{0.949,0.949,0.949}}34.85                     \\
ERASOR\textsuperscript{*}~\cite{9361109}            & 66.70                    & 98.54                    & {\cellcolor[rgb]{0.949,0.949,0.949}}81.07                     & 69.40                    & 99.06                    & {\cellcolor[rgb]{0.949,0.949,0.949}}82.92                     & 77.51                    & 99.18                    & {\cellcolor[rgb]{0.949,0.949,0.949}}\bf{87.68}                     & 94.90                    & 66.26                    & {\cellcolor[rgb]{0.949,0.949,0.949}}79.30                     \\
Octomap~\cite{hornung13auro}                        & 68.05                    & 99.69                    & {\cellcolor[rgb]{0.949,0.949,0.949}}82.37                     & 66.28                    & 99.24                    & {\cellcolor[rgb]{0.949,0.949,0.949}}81.10                     & 65.91                    & 96.70                    & {\cellcolor[rgb]{0.949,0.949,0.949}}79.84                     & 88.97                    & 82.18                    & {\cellcolor[rgb]{0.949,0.949,0.949}}\bf{85.51}                     \\
Octomap w G                                         & 85.92                    & 98.88                    & {\cellcolor[rgb]{0.949,0.949,0.949}}92.17                     & 86.15                    & 98.46                    & {\cellcolor[rgb]{0.949,0.949,0.949}}92.10                     & 76.38                    & 86.26                    & {\cellcolor[rgb]{0.949,0.949,0.949}}81.17                     & 94.95                    & 73.95                    & {\cellcolor[rgb]{0.949,0.949,0.949}}83.80                     \\
Octomap w GF                                        & 93.06                    & 98.67                    & {\cellcolor[rgb]{0.949,0.949,0.949}}\bf{95.83}                     & 93.54                    & 92.48                    & {\cellcolor[rgb]{0.949,0.949,0.949}}\bf{93.01}                     & 82.66                    & 82.44                    & {\cellcolor[rgb]{0.949,0.949,0.949}}82.55                     & 96.79                    & 73.50                    & {\cellcolor[rgb]{0.949,0.949,0.949}}84.34                     \\

\bottomrule
\end{tabular}
\label{num_table}
\end{table*}
\renewcommand{\arraystretch}{1.0}
\subsection{Metric}
Although nowadays datasets provide point-wise labels, most methods downsample the ground truth to voxel-wise level for evaluation. To provide a more accurate evaluation, we propose a new benchmark based on point-wise assessments.

The map clean task aims for two goals: remove true dynamic points and keep true static points. This process involves maintaining high recall in the classification of both dynamic and static points, often referred to as Dynamic Accuracy (DA\%) and Static Accuracy (SA\%) respectively.

Additionally, we utilize the Associated Accuracy (AA \%) calculated using the geometric mean as a comprehensive metric that combines both accuracies, offering an overall assessment of the algorithm's performance. AA is more sensitive to smaller values compared to the harmonic mean used in the F1 score.

\begin{align}
    AA = \sqrt{SA \times DA}.
\end{align}

There are also distance distribution plots that show the distance from mislabeled points to their nearest correct dynamic points. It serves as a metric to illustrate where errors typically occur, helping researchers identify and address the shortcomings of methods for further improvements.

\subsection{Implementation details}
We conduct experiments on three primary datasets: KITTI, Argoverse 2.0, and a semi-indoor dataset. The first two have their own ground truth pose files, while semi-indoor dataset poses are obtained using the SLAM package~\cite{simple-ndt-slam}. All datasets are integrated into a unified PCD format with point cloud data and pose in it.
KITTI has ground truth labels for dynamic objects from SemanticKITTI~\cite{behley2019iccv}. Part of Argoverse 2.0 and the semi-indoor dataset collected by us have manually labeled dynamic points ground truth. In the point-wise evaluation, if an algorithm rasterizes the grid, we query all the points in ground truth to search for the corresponding grid and label the point as static or dynamic according to the algorithm's output.

\section{Benchmark Results}
\label{sec:evaluation}

In this section, our benchmark contains both quantitative and qualitative evaluations. We use it to conduct a detailed analysis of the performance of the methods in Section~\ref{sec:benchmark_methods} on various datasets, identify specific failure scenarios for each method, and explain the reasons for these failures in relation to their theoretical foundations. Additionally, we provide a table outlining the time cost and the number of parameters required for tuning to achieve a better static map.

All experiments are conducted on a desktop computer equipped with a 12th Gen Intel® Core™ i9-12900KF processor featuring 24 cores.
The benchmark link includes all the parameters used for the experiments presented in this paper. 
Methods marked with an asterisk (*) in tables and figures indicate offline methods, which require a raw global point cloud map as a prior for comparison. More detail can be found in Section~\ref{sec:benchmark_methods} and Fig.~\ref{fig:framework}.

\subsection{Quantitative}
\begin{figure}[t]
    \centering
    \includegraphics[width=0.5\textwidth]{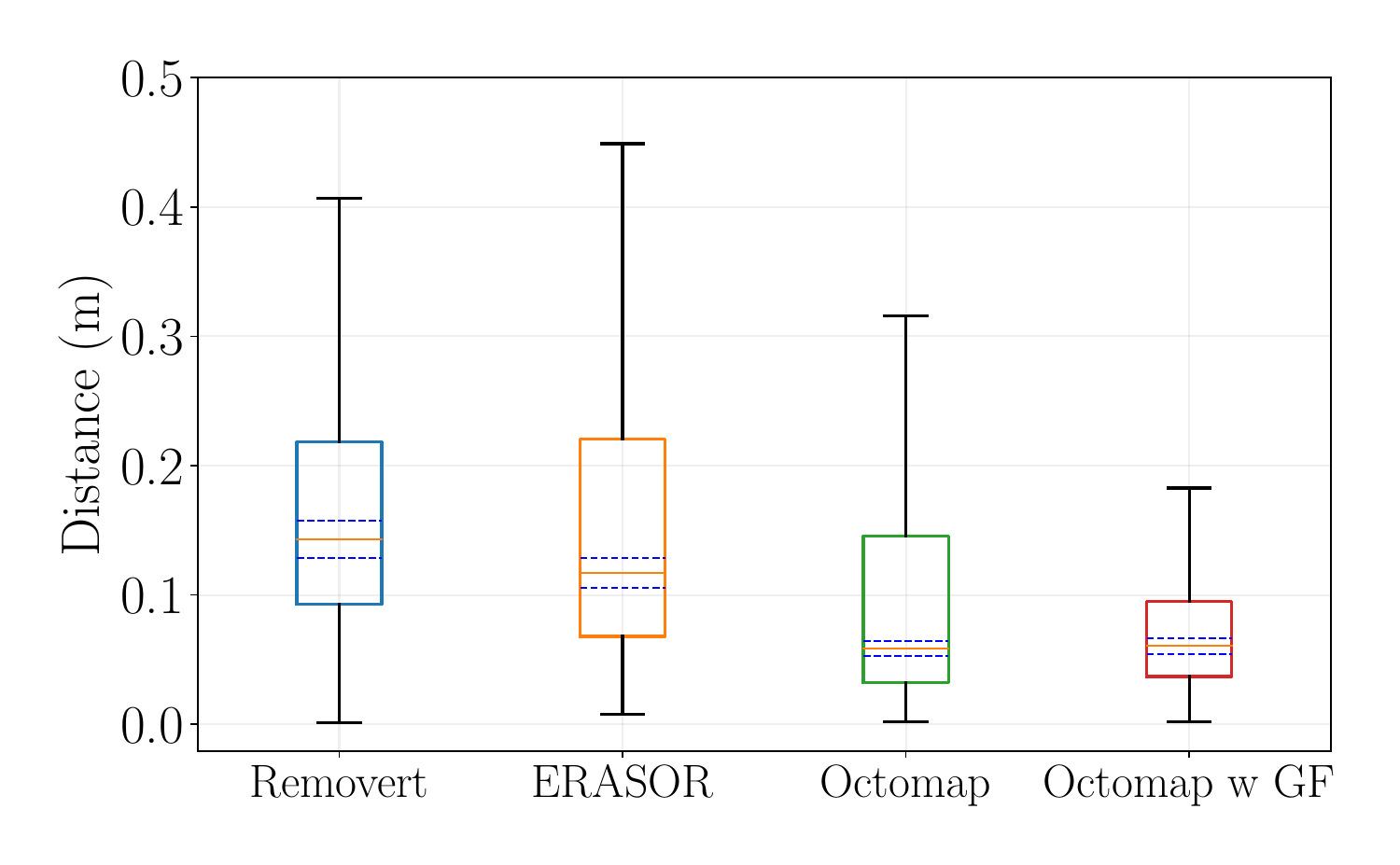}
    \caption{The distribution of distances from false negative points to their nearest true positive points for different algorithms in KITTI sequence 05. 
    The blue dashed line signifies the region containing 10\% of the point numbers.}
    \label{fig:distance_distribution}
    \vspace{-1.0em} 
\end{figure}

In Table \ref{num_table}, we present a quantitative comparison of dynamic object removal methods in various scenarios datasets. Removert mostly retains the complete static map but labels only a few correct dynamic points. In contrast, ERASOR performs better in removing dynamic points and balancing static and dynamic points. The original Octomap suffers from angle problems in the ground plane and noise points shown in Section~\ref{sec:qual} which cause their score on SA to be lower than others. Octomap w G denotes the method with ground estimation, resulting in a higher score across most of the sequences. 
Also adding the noise filter (Octomap w GF), we achieve a $20\%-30\%$ speed-up, as shown in Table \ref{time_table}, since the noise points do not undergo the ray casting process.

Considering Table \ref{time_table} and Table \ref{num_table}, there is a trade-off between speed and performance, as the method with the lowest score on AA achieves the fastest processing time for a single frame. It is important to note that Removert requires multiple resolutions to produce better results, which may increase the processing time depending on the number of resolutions. The speed of ray-based methods like Octomap has the potential for further optimization and improvement.

To better future analyze, Fig.~\ref{fig:distance_distribution} illustrates where errors typically occur. 
We observe that most of the false negative points (true dynamic points labeled as static) are close to the true positive points (correctly labeled dynamic points), with all methods' false negative points ranging from $10\text{cm}$ to $30\text{cm}$ away from the true positive points. 
In such cases, other techniques, such as clustering around the true positive points, can be employed to address this issue. 
The largest scale difference occurs in Removert, corresponding to the challenges we mentioned earlier in visibility-based methods that involve occlusion behind the true positive points. Techniques that apply understanding object relationships and establishing connections between them may help address this issue more effectively.

\renewcommand{\arraystretch}{1.2}
\begin{table}[t]
\centering
\caption{Runtime comparison of different methods} 
\label{tab:runtime_comparison}
\begin{tabular}{l c c}
\toprule
Methods      & Runtime/frame [s] & \# Parameters\\
\midrule
Removert\textsuperscript{*}~\cite{gskim-2020-iros} & \bf{0.044 $\pm$ 0.002}  & 6      \\
ERASOR\textsuperscript{*}~\cite{9361109}   & 0.718 $\pm$ 0.039     & 18   \\
Octomap~\cite{hornung13auro}      & 2.985 $\pm$ 0.961    & \bf{5}    \\
Octomap w G  & 3.054 $\pm$ 0.966 & 8       \\
Octomap w GF & 2.147 $\pm$ 0.468 & 10      \\
\bottomrule
\end{tabular}
\label{time_table}
\vspace{-0.5em}
\end{table}
\renewcommand{\arraystretch}{1.0}

\begin{figure*}[ht]
  \centering
  
  \includegraphics[width=1.0\linewidth]{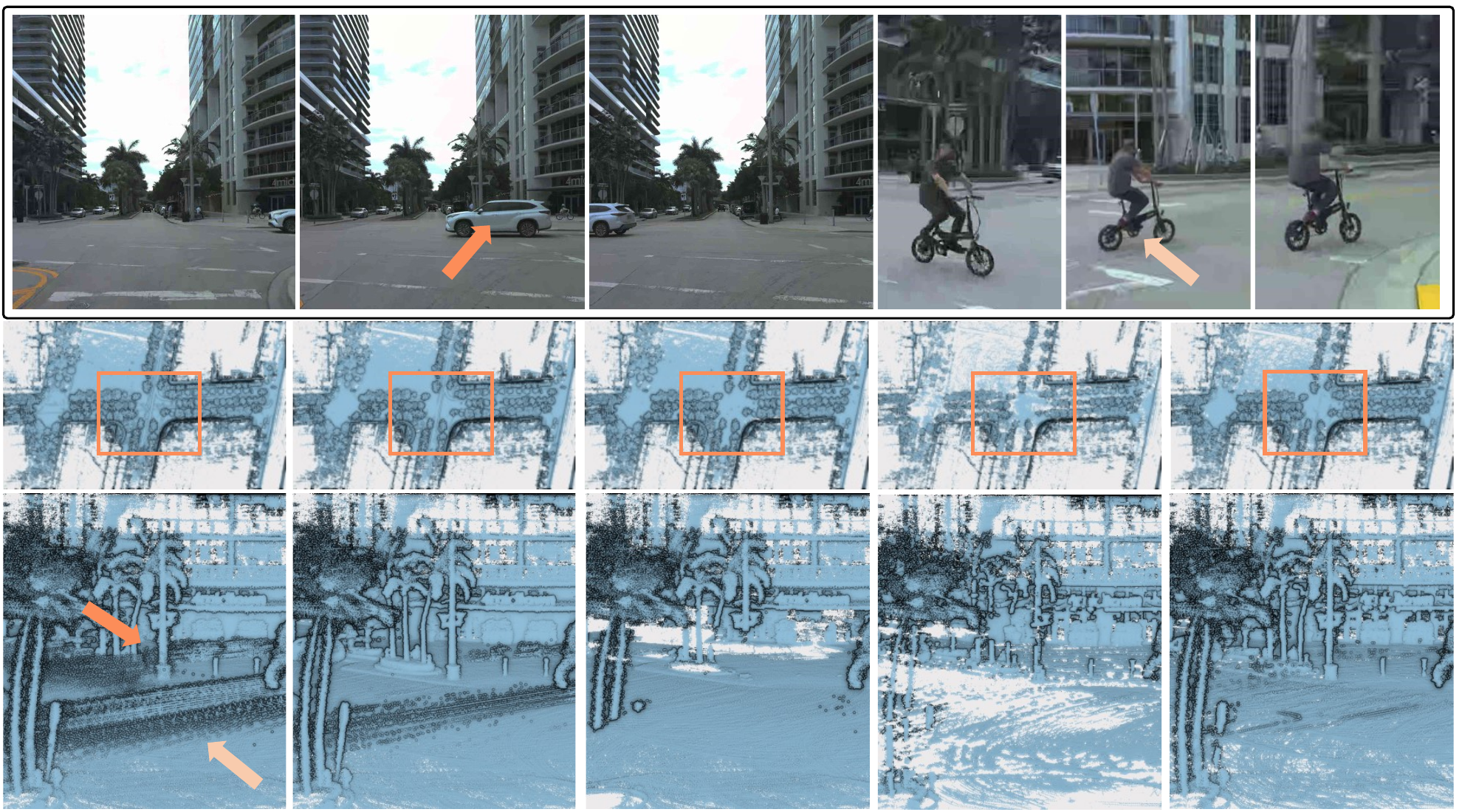}
  \newcolumntype{C}{>{\centering\arraybackslash}X}
\begin{tabularx}{\linewidth}{@{} *{5}{C} @{}}
  \small (a) Raw Map & \small (b) Removert\textsuperscript{*}~\cite{gskim-2020-iros} & \small (c) ERASOR\textsuperscript{*}~\cite{9361109} & \small \hspace{-15pt}(d) Octomap~\cite{hornung13auro} & \small \hspace{-5pt}(e) Octomap w GF \\[6pt]
\end{tabularx}
\vspace{-10pt}
\caption{\small Qualitative results for the Argoverse 2.0 big city with two VLP-32C LiDARs. The first row displays camera recordings from the cars, the middle row presents the entire map after methods cleaned for this sequence, and the bottom row focuses on a more detailed portion of the map corresponding to the image where cars and cyclists are present in the scene.}

  \label{fig:av2}
\end{figure*}

\subsection{Qualitative} 
\label{sec:qual}
To complement the quantitative results discussed earlier, We present the cleaned map in the Argoverse 2.0 and semi-indoor dataset, where the ground truth map is marked with yellow points to represent dynamic objects. 

In one sequence of the Argoverse 2.0 LiDAR dataset, Fig.~\ref{fig:av2} presents the cleaned maps produced by different methods in the Argoverse 2.0 dataset. 
As this dataset contains more recent and challenging scenarios from various US cities, it features many poles and trees that effectively illustrate the disadvantages of each method. 
As seen in the raw map, there are dynamic cars, cyclists, and pedestrians near the building. Removert retains the most complete static points but fails to remove the points near the object center. ERASOR keeps the cleanest map among all methods but, removes tree trunks and the ground near the pedestrian due to its sensitivity to height and slightly different pavement heights compared to driving roads. 
A comparison with the improved Octomap version in Fig.~\ref{fig:av2} (d) and  Fig.~\ref{fig:av2} (e) demonstrates significant improvements in error reduction for ground points. 
By incorporating ground estimation, most ground points are preserved, ensuring a more accurate representation of the static environment. As these points are considered definitively static, ray casting is not performed to remove ground points, further enhancing the accuracy of the map.
There is room for further improvement by using better ground estimation and clustering techniques to label the missing dynamic points in the map, as discussed in Fig.~\ref{fig:distance_distribution}.

\begin{figure*}[ht]
  \centering
  \includegraphics[width=1.0\linewidth]{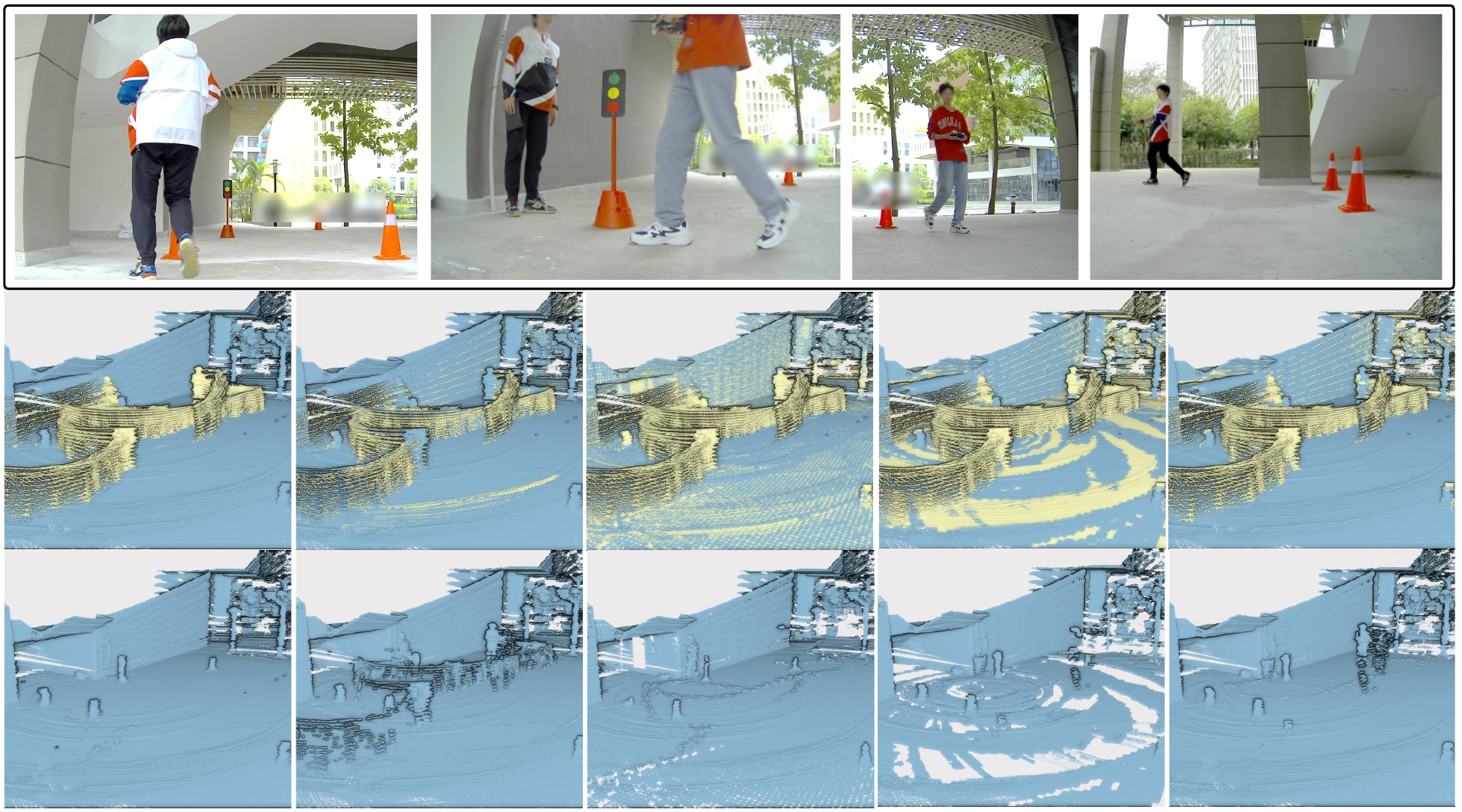}
  \newcolumntype{C}{>{\centering\arraybackslash}X}
\begin{tabularx}{\linewidth}{@{} *{5}{C} @{}}
  \small (a) Ground Truth & \small (b) Removert\textsuperscript{*}~\cite{gskim-2020-iros} & \small (c) ERASOR\textsuperscript{*}~\cite{9361109} & \small \hspace{-15pt}(d) Octomap~\cite{hornung13auro} & \small \hspace{-5pt}(e) Octomap w GF \\[6pt]
\end{tabularx}
\vspace{-10pt}
\caption{\small Qualitative results for a custom dataset with one sparse VLP-16 LiDAR. \colorbox[HTML]{FFFFBF}{\textcolor{black}{Yellow}} indicates points predicted as dynamic by the algorithm, while the first column shows ground truth labels provided by human annotation.}

  \label{fig:vlp16}
\end{figure*}
Fig.~\ref{fig:vlp16} displays the cleaned map for a custom dataset with one VLP-16 LiDAR. As clearly illustrated, the issues discussed in Section \ref{sec:related_work_traditional} are apparent. Removert has difficulty removing points that are behind dynamic obstacles moving around. Without fine-tuning the parameters and using the same settings as for the KITTI dataset, ERASOR fails to remove points higher than the threshold, and the original Octomap exhibits the sparse LiDAR ground problem, leading to many ground points being removed regularly by LiDAR rings. 
The improved Octomap still requires some fine-tuning of the occupancy probability values, as people are standing in the same place for an extended period, making it challenging to remove them using the default parameters.



\section{Conclusion}
In this paper, we conducted a comprehensive review and benchmark of methods for removing dynamic points from point clouds. We refactored three existing methods and contributed them to a unified benchmarking framework. The proposed metric on error distribution offers a novel perspective for analyzing where errors occur and gaining insights for researchers. 

In benchmarking evaluation, we provide detailed analyses of each method's strengths and weaknesses for future researchers. 
Guiding by our evaluation, we also propose a modified Octomap version tailored for this task by filtering and estimating ground points first. Through analysis, there is potential to generalize methods to similar scenarios that minimize reliance on parameter tuning and prior knowledge, as well as accelerate the algorithms for efficient execution. 

The future direction of this benchmark extends beyond merely removing dynamic points to encompass generating labels in perception datasets, as demonstrated in studies such as \cite{chen2022ral}, or performing real-time detection in point clouds, as illustrated by \cite{schmid2023dynablox}.

In conclusion, we hope this benchmark, open-source code, and dataset will serve as valuable resources for researchers and practitioners in this field, fostering further advancements and innovations in point cloud processing.

\section*{Acknowledgement}
Thanks to RPL's members: Yi Yang, and HKUST Ramlab's members: Bowen Yang, Jin Wu, and Yingbing Chen, who gave constructive comments on this work. 
Thanks to Shenzhen Unity-Drive Inc. for providing essential experimental devices and services for this work.
We also thank the anonymous reviewers for their constructive comments.

This work was partially supported by the Wallenberg AI, Autonomous Systems and Software Program (WASP) funded by the Knut and Alice Wallenberg Foundation.

\bibliographystyle{IEEEtran}
\bibliography{IEEEabrv,ref}

\end{document}